\documentclass[10pt,letterpaper]{article}
\usepackage{aaai20}
\usepackage{times}
\usepackage{helvet}
\usepackage{courier}
\nocopyright
\usepackage{tabularx}
\usepackage{xparse}
\usepackage{amsmath}
\usepackage{amsthm}
\usepackage{amssymb}
\usepackage{xspace}
\usepackage{relsize}
\usepackage{graphicx}
\usepackage{multirow}
\usepackage{url}
\usepackage{comment}
\usepackage{math_commands}
\usepackage{adjustbox}
\usepackage{algorithm}
\usepackage{algorithmicx}
\usepackage[noend]{algpseudocode}

\usepackage[frozencache]{minted}

\usepackage{xr}

\newcommand{\subsubparagraph}[1]{}

\makeatletter
\let\@myref\ref

\newcommand{\refsec}[1]{Sec.\,\@myref{#1}}
\newcommand{\refseq}[1]{Sec.\,\@myref{#1}}
\newcommand{\refig}[1]{Fig.\,\@myref{#1}}
\newcommand{\refigs}[2]{Fig.\,\@myref{#1}-\@myref{#2}}

\newcommand{\reftbl}[1]{Table \@myref{#1}}
\newcommand{\refstep}[1]{Step \@myref{#1}}
\newcommand{\refalgo}[1]{Algorithm \@myref{#1}}
\newcommand{\refchap}[1]{Chapter \@myref{#1}}
\newcommand{\reflst}[1]{List \@myref{#1}}
\newcommand{\refeq}[1]{Eq. \@myref{#1}}

\makeatother

\newcounter{list}[section]

\newcommand{\brackets}[1]{{\left<#1\right>}}
\newcommand{\braces}[1]{{\left\{#1\right\}}}
\newcommand{\parens}[1]{{\left(#1\right)}}

\newcommand{\sota}{State-of-the-Art\xspace}
\newcommand{\lsota}{state-of-the-art\xspace}  

\newcommand{\astar}{\xspace {$A^*$}\xspace}

\def\_{\\[-0.3em]}

\makeatletter

\newcommand{\newheuristic}[2]{%
 \def#1{%
  \ifmmode%
  h^\text{#2}\xspace%
  \else%
  \text{#2}\xspace%
  \fi%
 }%
}

\newheuristic{\lmcut}{LMcut}
\newheuristic{\mands}{M\&S}
\newheuristic{\pdb}{PDB}
\newheuristic{\ff}{FF}
\newheuristic{\ce}{CEA}
\newheuristic{\cg}{CG}
\newheuristic{\ad}{add}
\newheuristic{\lc}{LC}
\newheuristic{\hmax}{max}

\newcommand{\newUnitCostHeuristic}[2]{%
 \def#1{%
  \ifmmode%
  \hat{h}^\text{#2}\xspace%
  \else%
  \text{#2}\xspace%
  \fi%
 }%
}

\newUnitCostHeuristic{\lmcuto}{LMcut}
\newUnitCostHeuristic{\mandso}{M\&S}
\newUnitCostHeuristic{\ffo}{FF}
\newUnitCostHeuristic{\ceo}{CEA}
\newUnitCostHeuristic{\cgo}{CG}
\newUnitCostHeuristic{\ado}{add}
\newUnitCostHeuristic{\gco}{GoalCount}
\newUnitCostHeuristic{\lco}{LC}

\makeatother

\def\latentplanner{Latplan\xspace}

\newcommand{\var}[1]{\textsf{\relsize{-1}#1}}
\newcommand{\function}[1]{\textsc{#1}}

\newcommand{\init}{{\vx^{I}}}
\newcommand{\goal}{{\vx^{G}}}
\newcommand{\zinit}{{\vz^{I}}}
\newcommand{\zgoal}{{\vz^{G}}}
\newcommand{\encode}{\function{Encode}}
\newcommand{\decode}{\function{Decode}}

\newcommand{\Tr}{\text{Tr}}

\newcommand{\defaultindex}{i}
\newcommand{\tr}[1][\defaultindex]{\text{tr}^{#1}}
\newcommand{\xbefore}[1][\defaultindex,0]{\vx^{#1}}
\newcommand{\xafter}[1][\defaultindex,1]{\vx^{#1}}
\newcommand{\zbefore}[1][\defaultindex,0]{\vz^{#1}}
\newcommand{\zafter}[1][\defaultindex,1]{\vz^{#1}}
\newcommand{\zafterrec}[1][\defaultindex,1]{\tilde{\vz}^{#1}}
\newcommand{\action}[1][\defaultindex]{\va^{#1}}

\newcommand{\aaee}{\function{action}}
\newcommand{\aaed}{\function{apply}}

\newcommand{\len}[1]{|#1|}

\newcommand{\arr}[1]{\parens{#1}}

\newcommand{\pre}[1]{\function{pre}\parens{#1}}
\newcommand{\adde}[1]{\function{eff}^+\parens{#1}}
\newcommand{\dele}[1]{\function{eff}^-\parens{#1}}

\def\ref{\todo{Do not use ``ref'' directly!}}

\author{Masataro Asai \\ MIT-IBM Watson AI Lab, Cambridge USA \\ IBM Research}
\title{Neural-Symbolic Descriptive Action Model from Images: \\ The Search for STRIPS}

\usepackage{fancyhdr}
\begin{document}
\thispagestyle{fancy}
\maketitle

\begin{abstract}
Recent work on Neural-Symbolic systems that learn the discrete planning model from images has
opened a promising direction for expanding the scope of Automated Planning and Scheduling to the raw, noisy data.
However, previous work only partially addressed this problem,
utilizing the black-box neural model as the successor generator.
In this work, we propose Double-Stage Action Model Acquisition (DSAMA),
a system that obtains a descriptive PDDL action model with explicit preconditions and effects
over the propositional variables unsupervised-learned from images.
DSAMA trains a set of Random Forest rule-based classifiers and compiles them into logical formulae in PDDL.
While we obtained a competitively accurate PDDL model compared to a black-box model,
we observed that the resulting PDDL is too large and complex for the \lsota standard planners such as Fast Downward
primarily due to the PDDL-SAS+ translator bottleneck.
From this negative result, we show that
this translator bottleneck \emph{cannot} be addressed
just by using a different, existing rule-based learning method,
and we point to the potential future directions.
\end{abstract}

\section{Introduction}
\label{sec:introduction}

Recently, Latplan system \cite{Asai2018} successfully connected a
subsymbolic neural network (NN) system and a symbolic Classical Planning
system to solve various visually presented puzzle domains.
The system consists of four parts:
1) The State AutoEncoder (SAE) neural network
learns a bidirectional mapping between images and
propositional states with unsupervised training.
2) Action Model Acquisition module generates an action model from
the propositional state transitions encoded from the images.
3) Classical Planning module solves the problem in the
propositional state space with the learned action model.
4) The decoding module maps the propositional plan back to an image sequence.
The proposed framework opened a promising direction for applying a variety of symbolic
methods to the real world ---
For example, the search space generated by \latentplanner
was shown to be compatible with a symbolic Goal Recognition system
\cite{amado2018goal,amado2018goalb}.
Several variations replacing the state encoding modules have also been proposed:
Causal InfoGAN \cite{kurutach2018learning} uses a GAN-based framework,
First-Order SAE \cite{Asai2019b} learns the First Order Logic symbols (instead of the propositional ones),
and Zero-Suppressed SAE \cite[ZSAE]{Asai2019a} addresses the Symbol Stability issue of the regular SAE with $\ell_1$ regularization.

\begin{figure}[tb]
 \centering
 \relsize{-2}
 \begin{minted}{common-lisp}
(:action a0 :parameters () :precondition [D0]
 :effect (and (when [E00]            (z0))
              (when (not [E00]) (not (z0)))
              (when [E01]            (z1))
              (when (not [E01]) (not (z1))) ...))
 \end{minted}
 \caption{An example DSAMA compilation result for the first action (i.e. \texttt{a0})
generated from the image planning domain. \texttt{[...]} consists of a negation normal form compiled from a Random Forest \cite[RF]{ho1998random}.} %
 \label{rf-example}
\end{figure}

Despite these efforts,
Latplan is missing a critical feature of the traditional
Classical Planning systems: The use of \sota heuristic functions.
The main reason behind this limitation is the lack of \emph{descriptive} action model consisting of
logical formula for the preconditions and the effects,
which allows the heuristics to exploit its causal structures.
\emph{Obtaining the descriptive action models from the raw observations with minimal human interference} is
the next key milestone for expanding the Automated Planning applications to the raw unstructured inputs,
as it fully unleashes the pruning power of \lsota Classical Planning heuristic functions
which allow the planner to scale up to much larger problems.

In this paper, we propose an approach
called Dual-Stage Action Model Acquisition (DSAMA), a dual-stage process that
first learns the set of action symbols and action effects via Action
AutoEncoder neural network module in Latplan AMA$_2$ \cite{Asai2018} and
then trains a rule-based machine learning system
which are then converted into propositional formula in a PDDL format.
We tested DSAMA with 
Random Forest (RF) framework \cite{ho1998random} as the machine learning module due to its maturity and performance.
As a result, we successfully generated a descriptive action model, as depicted in \refig{rf-example} for example,
which is as accurate as the black-box neural counterpart.

Despite the success in terms of the model accuracy,
the proposed approach turned out to be an impractical solution for descriptive action model acquisition
and gave us an insight into the core problem of this approach.
The generated logical formula and the resulting PDDL was too large and complex
for the recipient classical planning system (Fast Downward)
to solve the given instance in a reasonable runtime and memory,
and if we trade the accuracy with the model simplicity, the goal becomes unreachable.
We provide an analysis on the reason and discuss possible future directions.
The code reproducing the experiments will be published at \texttt{github.com/guicho271828/latplan/}.

\section{Preliminaries}
\label{preliminary}

We denote a tensor (multi-dimensional array) in bold and denote its elements
with a subscript, e.g. when $\vx\in \R^{N\times M}$, the second row is $\vx_2 \in \R^M$.
We use dotted subscripts to denote a subarray, e.g. $\vx_{2..5}=(\vx_2,\vx_3,\vx_4)$.
For a vector or a set $X$, $|X|$ denotes the number of elements.
$\1^D$ and $\0^D$ denote the constant matrix of shape $D$ with all elements being 1/0, respectively.
$\va;\vb$ denotes a concatenation of tensors $\va$ and $\vb$ in the first axis
where the rest of the dimensions are same between $\va$ and $\vb$.
For a dataset, we generally denote its $i$-th data point with a superscript $^{i}$
which we may sometimes omit for clarity.

Let $\mathcal{F}(V)$ be a propositional formula consisting of
 logical operations $\braces{\land,\lor,\lnot}$,
 constants $\braces{\top,\bot}$, and
 a set of propositional variables $V$.
We define a grounded (propositional) Classical Planning problem
as a 4-tuple $\brackets{P,A,I,G}$
where
 $P$ is a set of propositions,
 $A$ is a set of actions,
 $I\subset P$ is the initial state, and
 $G\subset P$ is a goal condition.
Each action $a\in A$ is a 3-tuple $\brackets{\pre{a},\adde{a},\dele{a}}$ where
 $\pre{a} \in \mathcal{F}(P)$ is a precondition and
 $\adde{a}$, $\dele{a}$ are the sets of effects called add-effects and delete-effects, respectively.
Each effect is denoted as $c \triangleright e$ where
 $c \in \mathcal{F}(P)$ is an \emph{effect condition} and
 $e \in P$.
A state $s\subseteq P$ is a set of true propositions,
an action $a$ is \emph{applicable} when $s \vDash \pre{a}$ ($s$ \emph{satisfies} $\pre{a}$),
and applying an action $a$ to $s$ yields a new successor state $a(s)$ which is
$a(s) = s \cup \braces{e \mid (c \triangleright e) \in \adde{a}, c\vDash s} \setminus \braces{e \mid (c \triangleright e) \in \dele{a}, c\vDash s}$.

Modern classical planners such as 
Fast Downward \cite{Helmert04} takes the PDDL \cite{McDermott00} input which
specifies the above planning problem,
and returns an action sequence that reaches the goal state from the initial state.
Recent planners typically convert a propositional planning model into
SAS+ \cite{backstrom1995complexity} format, upon which the disjunctions in the
action model must be eliminated by
moving the disjunctions to the root of the formula and
splitting the actions \cite{helmert2009concise}.

\subsubsection{Latplan}
\label{background}

\latentplanner \cite{Asai2018} is a framework for
\emph{domain-independent image-based classical planning}.
It learns the state representation as well as the transition rules
entirely from the image-based observation of the environment with deep neural networks
and solves the problem using a classical planner.

\latentplanner takes two inputs.
The first input is the \emph{transition input} $\Tr$, a set of pairs of raw data randomly sampled from the environment.
An $\defaultindex$-th data pair in the dataset $\tr=(\xbefore, \xafter) \in \Tr$ represents
a randomly sampled transition from an environment observation $\xbefore$ to another observation $\xafter$ where some unknown action took place.
The second input is the \emph{planning input} $(\init, \goal)$, a pair of raw data,
 which corresponds to the initial and the goal state of the environment.
The output of \latentplanner is a data sequence representing the plan execution
 $(\init,\ldots \goal)$ that reaches $\goal$ from $\init$.
While the original paper used an image-based implementation (``data'' = raw images),
the type of data is arbitrary as long as it is compatible with neural networks.

\latentplanner works in 3 steps.
In Step 1, a \emph{State AutoEncoder} (SAE) (\refig{three}, left) neural network learns a bidirectional mapping between raw data $\vx$ (e.g., images)
 and propositional states $\vz\in\braces{0,1}^F$, where the propositional states are represented by $F$-dimensional bit vector.
The network consists of two functions $\encode$ and $\decode$, where 
$\encode$ maps an image $\vx$ to $\vz=\encode(\vx)$, and $\decode$ function maps $\vz$ back to an image $\vy=\decode(\vz)$.
The training is performed by minimizing the reconstruction loss $||\vy-\vx||$ under some norm (e.g., Mean Square Error for images).
In order to guarantee that $\vz$ is a binary vector, the network must use a discrete latent representation learning method
such as Gumbel Softmax \cite{jang2016categorical,MaddisonMT17} or Step Function with straight-through estimator \cite{koul2018learning,bengio2013estimating}
 --- We used Gumbel Softmax annealing-based continuous relaxation
 $\function{GumbelSoftmax}(\vx)=\function{Softmax}((\vx-\log(-\log(\function{Uniform}(0,1))))/\tau)$, $\tau\rightarrow 0$ in this paper.
After learning the mapping from $\braces{\ldots\xbefore, \xafter\ldots}$,
SAE obtains the propositional transitions $\overline{\Tr}=$
$\braces{(\encode(\xbefore), \encode(\xafter))} = \braces{(\zbefore,\zafter)}$.
In Step 2, an Action Model Acquisition (AMA) method learns an action model from $\overline{\Tr}$.
In Step 3, a planning problem instance is generated from the planning input $(\init, \goal)$.
These are converted to the discrete states $(\zinit, \zgoal)$ and the classical planner finds the path connecting them.
For example, an 8-puzzle problem instance consists of an image of the start (scrambled) configuration of the puzzle and an image of the solved state.
In the final step, \latentplanner obtains a step-by-step, human-comprehensive visualization of the plan execution
by $\decode$'ing the latent bit vectors for each intermediate state
and validates the visualized result using a custom domain-specific validator, for the evaluation purpose.
This is because the SAE-generated latent bit vectors are learned unsupervised and not directly verifiable through human knowledge.

\subsubsection{Action Model Acquisition (AMA)}

\label{oldama}

\begin{figure*}[ht]
 \includegraphics[width=\linewidth]{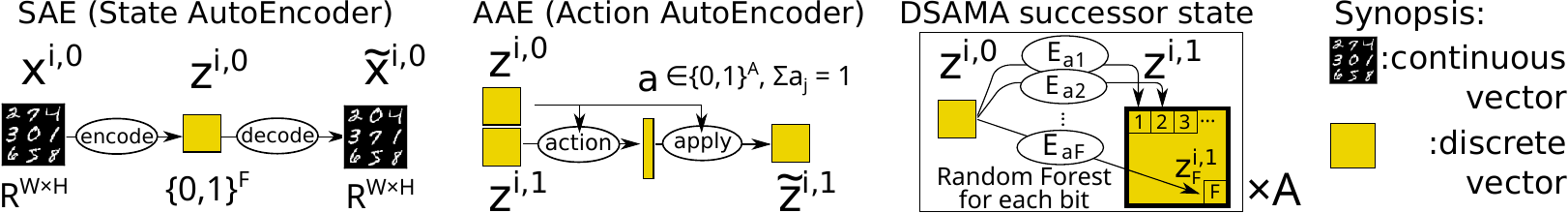}
 \caption{The illustration of State AutoEncoder, Action AutoEncoder, and Double Stage AMA for effect prediction.}
 \label{three}
\end{figure*}

The original Latplan paper proposed two approaches for AMA.
AMA$_1$ is an oracular model that directly generates a PDDL without learning,
and AMA$_2$ is a neural model that approximates AMA$_1$ by learning from examples.

AMA$_1$ is an oracular, idealistic AMA that does not incorporate machine learning,
and instead generates the entire propositional state transitions from the entire image transitions in the search space.
Each propositional transition is turned into a single, grounded action schema.
For example, in a state space represented by 2 latent space propositions $\vz=(\vz_1,\vz_2)$,
a transition from $\zbefore=(0,1)$ to $\zafter=(1,0)$ is translated into an action
with $\pre{a}= \lnot \vz_1\land \vz_2, \adde{a} = \braces{\top\triangleright\vz_1}, \dele{a} = \braces{\top\triangleright \vz_2}$.
It is impractical because it requires the entire image transitions, but also because
the size of the PDDL is proportional to the number of transitions in the state space,
slowing down the PDDL-SAS+ translation, preprocessing, and heuristic calculation at each search node.

AMA$_2$ consists of two neural networks: Action AutoEncoder (AAE) and Action Discriminator (AD).
AAE is an autoencoder that learns to cluster the state transitions into a (preset) finite number of action labels.
See \refig{three} (middle) for the illustration.

AAE's encoder takes a propositional state pair $(\zbefore,\zafter)$ as the input.
The last layer of the encoder is activated by a discrete activation function (such as Gumbel Softmax) to become a one-hot vector of $A$ categories,
 $\action\in\braces{0,1}^A$ ($\sum_{1\leq j \leq A} \action_j = 1$), where 
$A$ is a hyperparameter for the maximum number of action labels and $\action$ represents an action label.
For clarity, we use the one-hot vector $\action$ and the index $a^\defaultindex=\arg \max \action$ interchangeably.
AAE's decoder takes the current state $\zbefore$ and $\action$ as the input
and output $\zafterrec$, which is a reconstruction of $\zafter$.
The encoder $\aaee(\zbefore,\zafter)=\action$ acts as a function that tells ``what action has happened'' and
the decoder can be seen as a progression function $\aaed(\action,\zbefore)=\zafterrec$ (\refig{three}, middle).

AD is a binary classifier that models the preconditions of the actions.
AD learns the condition from the observed propositional state transitions $\overline{\Tr}$ and a ``fake'' state transitions.
Let $P=\overline{\Tr}$ and $U$ be the fake transitions.
$U$ could be generated by applying a random action $1 \leq a \leq A$ to the states in $\overline{\Tr}$.

This learning task is a Positive-Unlabeled learning task \cite[PU-learning]{elkan2008learning}.
While all examples in $P$ are guaranteed to be the positive (valid) examples obtained from the observations,
the examples in $U$ are \emph{unlabeled}, i.e., we cannot guarantee that the examples in $U$ are always negative (invalid).
Unlike the standard binary classification task, which takes the purely positive and the purely negative dataset,
PU-learning takes such a positive and an unlabeled dataset and returns a positive-negative classifier.
Under the assumption that the positive examples are i.i.d.-sampled from the entire distribution of positive examples,
one can obtain a positive-negative classifier $D(\vs)$ for the input $\vs$
by correcting the confidence value of a labeled-unlabeled classifier $D_{PU}(\vs)$ by an equation $D(\vs) = D_{PU}(\vs) / c(V_P)$,
where $V_P$ is a positive validation set and
$c(V_P)=\E_{\vs\in V_P} \left[D_{PU}(\vs)\right]$ is a constant computed after the training of $D_{PU}(\vs)$ \cite{elkan2008learning}.
In AD, $\vs$ is $(\zbefore;\zafter)$, i.e., the concatenation of the propositional current state and the successor state,
unlike the standard STRIPS setting where the precondition only sees the current state.

Combining AAE and AD yields a successor function that can be used for graph search algorithms:
It first enumerates the potential successor states from the current state $\zbefore[]$
by iterating $\zafter[a]=\textit{apply}(\zbefore[],\action[])$ over $1\leq a\leq A$,
then prunes the generated successor states using AD, i.e., whether $D(\zbefore[];\zafter[a])>0.5$.
The major drawback of this approach is that both AAE and AD are black-box neural networks,
and thus are incompatible with the standard PDDL-based planners and heuristics, and requires a custom heuristic graph search solver.

\section{Double-Stage Learning}

To overcome
the lack of
PDDL compatibility of the black-box NNs in AMA$_2$,
we propose Double-Stage Action Model Acquisition method (DSAMA) which consists of 3 steps:
(1) It trains the same AAE networks to identify actions and perform the clustering,
(2) transfers the knowledge to a set of Random Forest binary classifiers (\refig{three}, right), then finally
(3) converts the classifiers into logical preconditions / effects in PDDL.
Let $M$ be a process that returns a Random Forest binary classifier and
let $\function{toPDDL}(b)$ be a function that converts a classifier $b$ into a logical formula $\overline{b}$.
The overall DSAMA process is shown in \refalgo{dsama}.

\begin{algorithm}[htb]
\caption{ \emph{Double-Stage Action Model Acquisition} for $\overline{\Tr}$ with a Random Forest classifier $M$.}
\label{dsama}
\relsize{-1}
\begin{algorithmic}
 \State Perform AAE training on $\overline{\Tr}$ and obtain $\aaee$ function.
 \ForAll{$1\leq a \leq A$}
   \State $Z^0_a \gets \braces{\zbefore|(\zbefore,\zafter)\in\overline{\Tr}, \aaee(\zbefore,\zafter)=a}$
   \State $Z^1_a \gets \braces{\zafter |(\zbefore,\zafter)\in\overline{\Tr}, \aaee(\zbefore,\zafter)=a}$
   \State $Z_a \gets \braces{\zbefore;\zafter|(\zbefore,\zafter)\in\overline{\Tr}, \aaee(\zbefore,\zafter)=a}$
   \State $U_a   \gets \braces{\zbefore;\zafter|(\zbefore,\zafter)\in\overline{\Tr}, \aaee(\zbefore,\zafter)\not=a}$
 \ForAll{$1\leq f \leq F$}
   \State  $Z^1_{af} \gets \braces{\zbefore[]_f| \zbefore[] \in Z^1_a}$
   \State  $E_{af} \gets M(Z^0_a, Z^1_{af})$
   \State  $\overline{E}_{af} \gets \function{toPDDL}(E_{af})$ {\relsize{-1} (Effect conditions for $\vz_f$)}
 \EndFor
 \State $D_a \gets M((Z_a;U_a),\; (\1^{|Z_a|};\0^{|U_a|}))$ {\relsize{-2} ($(X;Y)=$ concat of $X$ and $Y$)}
 \State $\overline{D}_a \gets \function{toPDDL}(D_a)$ {\relsize{-1} (precondition)}
 \State \textbf{collect} action $\brackets{\overline{D}_a,\ \bigcup_f \braces{\overline{E}_{af} \triangleright \vz_f},\ \bigcup_f \braces{\lnot \overline{E}_{af} \triangleright \lnot \vz_f}}$.
 \EndFor
\end{algorithmic}
\end{algorithm}

In order to learn the action preconditions,
DSAMA performs a PU-learning
following Action Discriminator (\refsec{oldama}).
Similar to AD, it takes both the current and successor states as the input -- in the later experiments,
we show that the accuracy drops when we limit the input to the current state.
Unlike AD in AMA$_2$, DSAMA trains a specific classifier for each action.
For the action effects,
DSAMA learns the effect condition $c$ of the conditional effect $c\triangleright e$ in PDDL (\refsec{preliminary}).
DSAMA iterates over every action $a$ and every bit $f$ and trains a binary classifier $E_{af}$ that translates to $c$.

\subsubsection{Random Forest (RF)}
\label{sec:rf}

While the binary classifier in DSAMA could be any binary classifier that could be converted to a logical formula,
we chose Random Forest \cite{ho1998random}, a machine learning method based on decision trees.
It constructs an ensemble of decision trees $\vt=\arr{\vt_1\ldots\vt_T}$
and averages the predictions returned by each tree $\vt_i$.
We do not describe the details of its training, which is beyond the scope of this paper.
It is one of the most widely used rule-based learning algorithms whose
implementations are available in various machine learning packages \footnote{We used \texttt{cl-random-forest} \cite{masatoi}}.
To address the potential data imbalance, we used a Balanced Random Forest algorithm \cite{chen2004using}.

A decision tree for classification consists of decision nodes and leaf nodes.
A decision node is a 4-tuple $(i,\theta,\textit{left},\textit{right})$,
where each element is the feature index, a threshold, and
the left / right child nodes.
A leaf node contains a class probability vector $\vp\in [0,1]^C$, where $C$
is a number of classes to be classified, which is 2 in our binary case.
To classify a feature vector $\vx\in \R^F$, where $F$ is the number of
features, it tests $\vx_i \leq \theta$ at each decision node and recurses
into the left/right children depending on success.
When a single tree is used for classification, it takes the $\argmax$ over the probability vector at the leaf node
and returns the result as the predicted classification.
For an ensemble of decision trees, classification is performed either by
taking the average of $\vp$ returned by the trees and then taking an $\argmax$ over the classes,
or by taking the $\argmax$ at each leaf node and returning the majority class voted by each tree.

Since STRIPS/PDDL cannot directly represent the numeric averaging operation,
we simulate the voting-based inference of Random Forest in the PDDL framework by
compiling the RF into a \emph{majority gate} boolean circuit.

First, converting a decision tree into a boolean logic formula is straightforward.
Since we assume a binary input dataset,
the decision nodes and the leaf nodes can be recursively converted into a Negation Normal Form as in the \function{toPDDL}(\var{tree} $t$) method (\refalgo{bitonic}).

\begin{figure}[tb]
 \centering
 \includegraphics{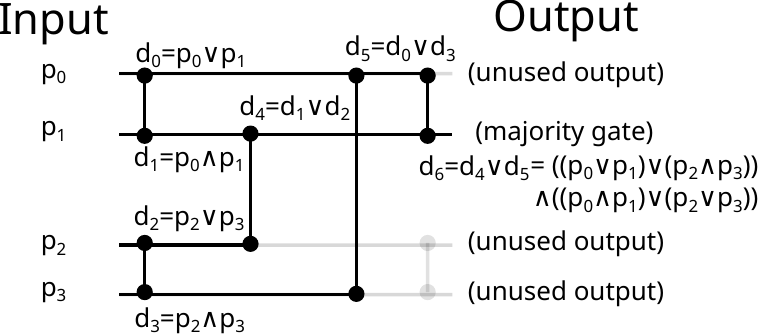}
 \caption{Majority-gate implementation for 4 inputs $p_0 \ldots p_3$, using 7 comparators $d_0 \ldots d_6$ generated as part of bitonic sorting network.}
 \label{bitonic-example}
\end{figure}

\begin{figure}[tb]
 \centering
 \includegraphics{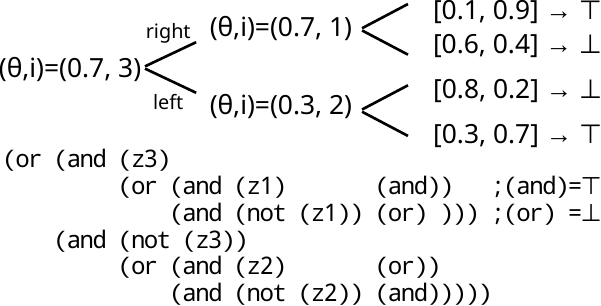}
 \caption{Compilation of a decision tree into a logical formula in PDDL (as a precondition or an effect condition).}
 \label{dtree-example}
\end{figure}

Next, recall that we now take the votes from all trees and
choose the class with the largest votes as the final
classification prediction. Since we are handling the binary classification,
finding the most voted class is equivalent to computing a \emph{majority function} \cite{lee1992bit},
a boolean function of $T$ fan-ins and a single fan-out
which returns 1 when more than $\lfloor T/2\rfloor$ inputs are 1.
One practical approach for implementing such a function is based on
\emph{bitonic sorting network} \cite{knuth1997art,batcher1968sorting}.
First, we apply the bitonic sorting algorithm proposed by \citeauthor{batcher1968sorting} (\refalgo{bitonic}) to $\vt_i$,
except that instead of taking the $\max$ and $\min$ in the \function{compareAndSwap} function,
we take $\lor$ and $\land$ of the elements being swapped
because $\max(x,y)=x\lor y$ and $\min(x,y)=x\land y$ for 0/1 values seen as boolean values.
We then use the $\lfloor T/2\rfloor$-th element stored in the result as the output.
See examples in \refig{bitonic-example}, \refig{dtree-example}, \refig{rf-example}.

Finally, since our preconditions $D_a$ takes the current and the successor states as the input,
we need to take care of the decision nodes that points to the successor state.
When the binary latent space has $F$ dimensions, the input vector to the random forest has $2F$ dimensions
where the first and the second half of $2F$ dimensions is for the current and the successor state.
In \function{toPDDL}, when the index $i$ of the decision node satisfies $F<i$,
we insert $\overline{E}_{a(i-F)}$ instead of $\vz_i$
because our DSAMA formulation guarantees that
$\vz_i$ is true in the successor bit when the effect condition $\overline{E}_{a(i-F)}$ is satisfied.
This can be seen as a trick to implement a one-step lookahead in the action model.

\begin{algorithm}[htb]
\caption{\function{toPDDL} method for Random Forest \cite{ho1998random} based on bitonic sorting network \cite{batcher1968sorting}}
\label{bitonic}
\relsize{-1}
\begin{algorithmic}
\Function{toPDDL}{\var{forest} $\vt=\parens{\vt_1\ldots\vt_T}$}
  \State $\function{sort}(\top,\vt)$ \Comment Bitonic sorting
  \State \Return $\vt_{\lfloor T / 2 \rfloor}$
\EndFunction
\Function{toPDDL}{\var{tree} $t$}
  \If{$t$ is a decision node $(i,\theta,\textit{left},\textit{right})$}
    \If{$0<\theta<1$}
    \State \Return $(\vx_i\land \function{toPDDL}(\textit{left})) \lor (\lnot \vx_i\land \function{toPDDL}(\textit{right}))$
    \EndIf
    \State \textbf{if} $\theta<0$ \Return $\function{toPDDL}(\textit{right})$, \textbf{else} \Return $\function{toPDDL}(\textit{left})$
  \ElsIf{$t$ is a leaf node $(p_0, p_1)$}
    \State \textbf{if} $p_0<p_1$ \Return $\top$, \textbf{else} \Return $\bot$
  \EndIf
\EndFunction
\Function{sort}{\var{bool} $u$, \var{vector} $\vx$}
  \If{$\len{\vx}\leq 1$} \Return $\vx$
  \Else \Comment Note: $(\va;\vb)$ is a concatenation of vector $\va$ and $\vb$ (\refsec{preliminary}).
    \State $d \gets \lfloor\len{\vx} / 2\rfloor,\quad \va \gets \vx_{0..d},\quad \vb \gets \vx_{d..\len{\vx}}$
    \State \Return $\function{merge}(u, (\function{sort}(\top, \va); \function{sort}(\bot, \vb))))$
  \EndIf
\EndFunction
\Function{merge}{\var{bool} $u$, \var{vector} $\vx$}
  \If{$\len{\vx}\leq 1$} \Return $\vx$
  \Else
    \State $\function{compareAndSwap}(u, \vx)$
    \State $d \gets \lfloor\len{\vx} / 2\rfloor,\quad \va \gets \vx_{0..d},\quad \vb \gets \vx_{d..\len{\vx}}$
    \State \Return $(\function{merge}(u, \va); \function{merge}(u, \vb))$
  \EndIf
\EndFunction
\Function{compareAndSwap}{\var{bool} $u$, \var{vector} $\vx$}
  \State $d \gets \lfloor\len{\vx} / 2\rfloor$
  \ForAll{$i \in \braces{0..d}$}
    \If{$u$} \Comment Compare in increasing order?
      \State $\vx_i     \gets \vx_i \lor  \vx_{i+d}$ \Comment originally $\max(\vx_i, \vx_{i+d})$
      \State $\vx_{i+d} \gets \vx_i \land \vx_{i+d}$ \Comment $\min(\vx_i, \vx_{i+d})$
    \Else                                     
      \State $\vx_i     \gets \vx_i \land \vx_{i+d}$ \Comment $\min(\vx_i, \vx_{i+d})$
      \State $\vx_{i+d} \gets \vx_i \lor  \vx_{i+d}$ \Comment $\max(\vx_i, \vx_{i+d})$
    \EndIf
  \EndFor
\EndFunction
\end{algorithmic}
\end{algorithm}

\section{Evaluation}

We evaluated our approach in the dataset used by \citeauthor{Asai2018}, which consists of 5 image-based domains.
\textbf{MNIST 8-puzzle}
is an image-based version of the 8-puzzle, where tiles contain hand-written digits (0-9) from the  MNIST database \cite{lecun1998gradient}.
Valid moves in this domain swap the ``0'' tile  with a neighboring tile, i.e., the ``0'' serves as the ``blank'' tile in the classic 8-puzzle. 
The \textbf{Scrambled Photograph 8-puzzle (Mandrill, Spider)} cuts and scrambles real photographs, similar to the puzzles sold in stores).
These differ from the MNIST 8-puzzle in that ``tiles'' are \textit{not} cleanly separated by black regions
(we re-emphasize that \latentplanner has no built-in notion of square or movable region).
\textbf{LightsOut} is
a video game where a 4x4 grid of lights is in some on/off configuration,
and pressing a light toggles its state as well as the states of its neighbors.
The goal is all lights Off.
\textbf{Twisted LightsOut} distorts the original LightsOut game image by a swirl effect, 
showing that \latentplanner is not limited to handling rectangular ``objects''/regions.
In all domains, we used 9000 transitions for training and 1000 transitions for testing.
Note that 8-puzzle contains 362880 states and 967680 transitions, and LightsOut contains 65536 states and 1048576 transitions.

We used the SAE with $F=100$, i.e., it produces 100 latent propositions.
Following the work of \cite{Asai2019a},
we used the standard version of SAE and a regularized version of SAE (ZSAE) with a regularization constant $\alpha=0.5$.
For the AAE, we tuned the upper-bound $A$ of the number of actions in AAE
by iteratively increasing $A$ from $A=8$ to $A=128$ by $\Delta A=8$
until the mean absolute error of AAE ($|\zafter-\zafterrec|$) goes below 0.01, i.e., below 1 bit on average.
This is because a large $A$ reduces the number of transitions that fall into a single action label
and makes the random forest training harder due to the lack of examples.

\subsection{Accuracy}

We compared the accuracy of DSAMA and AMA$_2$.
DSAMA has two primary controlling hyperparameters for Random Forest --- the maximum depth of the tree and the number of trees.
Other hyperparameters of Random Forest follows the standard parameters for classification tasks:
Entropy-based variable selection,
out-of-bag ratio 0.33 (each decision tree is trained on the random $\frac{2}{3}$ subset of the entire dataset),
and the number of variables considered by each tree as $\sqrt{F}$ for the $F$-dimensional dataset.
(note: this is not equivalent to the tree depth because the same variable may be selected several times.)

We first compared the successor generation accuracy between AAE (black-box NN model) and DSAMA.
The dataset $\overline{\Tr}$ is divided into the training set and the test set by 9:1.
AAE uses the same hyperparameters used in Latplan \cite{Asai2018}.
DSAMA uses a Random Forest with the number of trees $T=80$ and the maximum depth of the tree $D=100$,
the largest number we used in this paper. Note that, in general, Random Forest is claimed to achieve
the monotonically higher accuracy as it has more ensembles and depth.
\reftbl{accuracy-effect} shows the average reconstruction accuracy for the successor states over $F$ bits, over all transitions in the test dataset.
The results indicate that DSAMA based on Random Forest is competitive against the black box neural model.

Next, we compared
the F-measure based on 
the true positive rate (=\emph{recall}) and the true negative rate (=\emph{specificity})
of the black-box precondition model (Action Discriminator) and the DSAMA precondition model using Random Forest.
Note that this task is not only a PU-learning task, but also a classification task on a potentially highly imbalanced dataset
and therefore we cannot use the accuracy as the evaluation metric as it could be skewed toward the majority dataset \cite{wallace2011class}.

Similar results are obtained in \reftbl{accuracy-precondition}:
Rule-based method (Random-Forest) is competitive against the black box
method when a sufficiently large capacity is provided to the model
($T=80, D=100$).
To address the concern about using the successor states as part of the precondition,
we also tested the variants which learns only from the current state.
We observed a significant drop in the accuracy both in the black-box NN (AD) and the DSAMA $D_a$.

\begin{table}[tb]
\centering
\begin{adjustbox}{width={\linewidth},keepaspectratio}
\begin{tabular}{|llc||r|c||c|}
\multicolumn{3}{|c||}{SAE}   & \multicolumn{2}{c||}{AMA$_2$ AAE} & DSAMA $E_{af}$ \\
Domain   & $\alpha$ & MSE   &  $A$ & \multicolumn{2}{c|}{Successor bit accuracy} \\[0.2em]
LOut     & 0.0 & 2.25E-12 & 37 & 99.1\% & 98.5\% \\
LOut     & 0.5 & 8.18E-11 & 43 & 99.4\% & 99.2\% \\
Twisted  & 0.0 & 1.03E-02 & 48 & 99.3\% & 98.8\% \\
Twisted  & 0.5 & 1.08E-02 & 39 & 98.8\% & 98.4\% \\
Mandrill & 0.0 & 7.14E-03 & 14 & 99.9\% & 98.9\% \\
Mandrill & 0.5 & 7.09E-03 & 9  & 99.4\% & 98.9\% \\
Mnist    & 0.0 & 3.34E-04 & 67 & 98.8\% & 94.0\% \\
Mnist    & 0.5 & 3.44E-03 & 3  & 99.8\% & 99.5\% \\
Spider   & 0.0 & 9.51E-03 & 9  & 99.2\% & 98.5\% \\
Spider   & 0.5 & 8.88E-03 & 8  & 99.5\% & 98.4\% \\
\end{tabular}
\end{adjustbox}
 \caption{
Accuracy comparison between AAE and DSAMA on the successor state prediction task,
where the discrete transitions are generated by the SAE or ZSAE ($\alpha=0.5$) for 5 domains.
The accuracy is measured by the number of bits correctly predicted by the model in the test dataset.
DSAMA uses a Random Forest with the number of trees $T=80$ and the maximum depth of the tree $D=100$.
We observed that the Random Forest is competitive for this task against a black box NN and sometimes even outperformed it.
As a reference, we also showed:
(1) $A$, the number of action labels generated by the AAE, and
(2) the image reconstruction error of the SAE, where the values
are MSE for the pixel values in $[0,1]\subset \R$.
 }
\label{accuracy-effect}
\end{table}

\begin{table}[htb]
\begin{adjustbox}{width={\linewidth},keepaspectratio}
\begin{tabular}{|lr|c|c|c|c|}
         &          & \multicolumn{2}{c|}{AMA$_2$ AD} & \multicolumn{2}{c|}{DSAMA $D_a$}
 \\
\multicolumn{2}{|r|}{Input dataset $\rightarrow$}  & {$\overline{\Tr}$} & {$\braces{(\zbefore;\action)}$} & {$Z_a$} & {$Z^0_a$}
 \\[0.2em]
LOut     & $\alpha=$0.0 & \textbf{89.7\%} & 74.2\% & 81.5\%          & 75.0\% \\
LOut     & 0.5          & \textbf{89.1\%} & 68.7\% & 85.5\%          & 71.7\% \\
Twisted  & 0.0          & \textbf{84.3\%} & 75.5\% & 80.3\%          & 71.5\% \\
Twisted  & 0.5          & \textbf{84.5\%} & 72.6\% & 78.8\%          & 69.6\% \\
Mandrill & 0.0          & 81.6\%          & 76.2\% & \textbf{82.4\%} & 76.5\% \\
Mandrill & 0.5          & 66.0\%          & 29.1\% & \textbf{77.2\%} & 59.6\% \\
Mnist    & 0.0          & \textbf{87.5\%} & 81.6\% & 84.4\%          & 78.2\% \\
Mnist    & 0.5          & \textbf{63.8\%} & 57.0\% & 61.6\%          & 57.5\% \\
Spider   & 0.0          & \textbf{82.5\%} & 64.3\% & 80.7\%          & 69.2\% \\
Spider   & 0.5          & 60.1\%          & 15.9\% & \textbf{77.9\%} & 68.3\% \\
\end{tabular}
\end{adjustbox}
\caption{
F-measure of true positive rate (\emph{recall}) and true negative rate (\emph{specificity}),
comparing AD and DSAMA on the action applicability task.
The discrete transitions are generated by the SAE or ZSAE ($\alpha=0.5$) for 5 domains.
Given the prediction $X$ and the ground truth $G$,
$\text{recall}=P(X=1 \mid G=1)$, $\text{specificity}=P(X=0 \mid G=0)$, and
$\text{F}=\frac{2\cdot\text{recall}\cdot\text{specificity}}{\text{recall}+\text{specificity}}$.
The numbers are for a test dataset where each transition is assigned a boolean ground-truth value by a validator that works on the reconstructed image pairs.
DSAMA uses a Random Forest with the number of trees $T=80$ and the maximum depth of the tree $D=100$.
We show the results with the current state as the input, and those with the concatenation of the current and the successor states as the input.
The accuracy drops when the input dataset is limited to the current state only.
 }
\label{accuracy-precondition}
\end{table}

Next, in order to see the effect of the random forest hyperparameters on the learned results,
we performed an exhaustive experiment on $(T,D)\in \braces{1,2,5,10,20,40,80} \times \braces{4,7,12,25,50,100}$
and compared the precondition accuracy, the effect accuracy and the size of the PDDL files.
Note that $T=1$ is a degenerative case for a single decision tree without ensembles.
For the space constraint, we only show the results for Mandrill 8-Puzzle with ZSAE ($\alpha=0.5$),
but the overall characteristics were the same across domains and the choice of SAE / ZSAE.

We observed that the effect of larger $T$ and $D$ saturates quickly, while small numbers negatively affect the performance.
The action applicability prediction (i.e., the precondition accuracy, \reftbl{hyperparameters}, left) tends to be more affected by the depth $D$
while
the successor state reconstruction accuracy (i.e., the effect accuracy, \reftbl{hyperparameters}, middle) tends to be more affected by the number of trees $T$.
Larger $T$ and $D$ also implies larger file sizes.
(Note: When generating the PDDL file, we apply De-Morgan's law to simplify obvious invariants when one is encountered, e.g.,
$v \land \top = v$, $v \lor \bot = v$, $v \land \lnot v = \bot$ and $v \lor \lnot v = \top$.)

\begin{table*}[htb]
\begin{adjustbox}{width={\linewidth},keepaspectratio}
\begin{tabular}{|rr|}
                     &   \\
                     &   \\
                     &\  \\[0.2em]
\multirow{7}{*}{$T$} &1  \\
                     &2  \\
                     &5  \\
                     &10 \\
                     &20 \\
                     &40 \\
                     &80 \\
\end{tabular}
\begin{tabular}{rrrrrr|}
\multicolumn{6}{c|}{Action applicability (precondition) F-measure}\\
\multicolumn{6}{c|}{$D$}\\
4      & 7      & 12     & 25     & 50     & 100    \\[0.2em]
36.5\% & 56.1\% & 65.5\% & 49.2\% & 49.8\% & 49.5\% \\
40.5\% & 59.1\% & 67.0\% & 63.8\% & 63.8\% & 64.0\% \\
37.3\% & 56.8\% & 70.1\% & 69.0\% & 69.5\% & 68.8\% \\
35.7\% & 56.2\% & 71.1\% & 72.7\% & 72.9\% & 72.5\% \\
35.9\% & 56.8\% & 72.4\% & 75.0\% & 74.8\% & 74.6\% \\
32.8\% & 55.6\% & 73.1\% & 76.6\% & 76.2\% & 76.4\% \\
33.4\% & 55.6\% & 73.1\% & 77.1\% & 77.1\% & 77.2\% \\
\end{tabular}
\begin{tabular}{rrrrrr|}
\multicolumn{6}{c|}{Successor state reconstruction (effect) accuracy}\\
\multicolumn{6}{c|}{$D$}\\
4      & 7      & 12     & 25     & 50     & 100    \\[0.2em]
87.7\% & 90.6\% & 91.7\% & 91.8\% & 91.8\% & 91.8\% \\
91.5\% & 93.0\% & 91.8\% & 91.9\% & 91.9\% & 91.9\% \\
94.4\% & 95.8\% & 96.3\% & 96.3\% & 96.3\% & 96.3\% \\
95.5\% & 97.3\% & 97.5\% & 97.5\% & 97.5\% & 97.5\% \\
96.4\% & 98.0\% & 98.2\% & 98.2\% & 98.2\% & 98.2\% \\
96.7\% & 98.2\% & 98.8\% & 98.8\% & 98.8\% & 98.8\% \\
96.9\% & 98.5\% & 98.9\% & 98.8\% & 98.9\% & 98.9\% \\
\end{tabular}
\begin{tabular}{rrrrrr|}
\multicolumn{6}{c|}{PDDL domain file size}\\
\multicolumn{6}{c|}{$D$}\\
4    & 7    & 12   & 25   & 50   & 100  \\[0.2em]
964K & 1.5M & 2.0M & 3.0M & 2.9M & 2.9M \\ 
2.2M & 3.2M & 4.1M & 6.0M & 6.0M & 6.0M \\ 
7.8M & 11M  & 13M  & 18M  & 18M  & 18M  \\ 
24M  & 29M  & 34M  & 43M  & 44M  & 43M  \\ 
71M  & 81M  & 91M  & 109M & 109M & 109M \\ 
204M & 224M & 244M & 281M & 281M & 281M \\ 
557M & 597M & 636M & 710M & 710M & 709M \\
\end{tabular}
\end{adjustbox}
\caption{Accuracy for the precondition and the effects by DSAMA using various Random Forest hyperparameters,
 as well as the PDDL domain file sizes (in bytes) resulted from their compilation.
$T$ is the number of trees in each Random Forest ensemble, and $D$ is the maximum depth of the tree.
The table is showing the results for the test dataset of Mandrill 8-Puzzle, where the discrete state vectors are generated by ZSAE ($\alpha=0.5$).
 }
\label{hyperparameters}
\end{table*}

\subsection{Evaluation in the Latent Space}

To measure the effectiveness of our approach for planning,
we ran fast downward on the domain PDDL files generated by DSAMA system.

In each of the 5 domains, we generated 20 problem instances
$(\init,\goal)$ by generating the initial state with a random walk
from the goal state using a problem-specific simulator.
10 instances are generated with 7 steps away from the goal state while the others are generated with 14 steps.

We tested three scenarios: Blind search with \astar, FF
heuristics \cite{hoffmann01} with Greedy Best First Search, and
max-heuristics \cite{haslum2000admissible} with \astar.
We gave 1 hour time limit and a maximum of 256 GB memory to the planner.

We tested these configurations on a variety of PDDL domain files generated by different $T$ and $D$.
As stated in the introduction, despite our RF models achieving high accuracy in terms of prediction,
we did not manage to find a plan using Fast Downward.
The failure modes are threefold: The planner failed to find the goal after exhaustively searching the state space,
the initial heuristic value being infinity in the reachability analysis (in $\ff$ and $\hmax$),
or the problem transformation to SAS+ does not finish within the resource limit.

From the results in the previous tables, the reason of the failure is obvious:
There is a trade off between the accuracy and the PDDL file size.
When the PDDL model is inaccurate, the search graph becomes disconnected and the search fails.
If we increase the accuracy of the PDDL model, the file size increases and Fast Downward fails even to start the search.
Moreover, we observed the translation fails even with a PDDL domain file with the moderate file size (e.g. $(T,D)=(10,7)$, 11MB).

In order to narrow down the reason for failure, we tested the domain
files whose preconditions are removed, i.e., replaced with \texttt{(and)} and made always applicable.
We expected the planner to find any sequence of actions which may not be a valid solution.
The results were the same: The goal state is unreachable for the small PDDL files due to the lack of accuracy and
the translation does not finish for the large PDDL files.
Considering the fact that the effect of an action is modeled by $F$ random forests while the precondition is modeled by a single random forest,
we conclude that the effect modeling is the main bottleneck of the translator failure.
Note that, however, the maximum accuracy of the effect modeling with DSAMA is comparable to the neural model and quite high (typically $>98\%$).
We analyze this phenomenon in the next section.

\section{Discussion}

Our experiments showed that Random-Forest based DSAMA approach
does not work even if it achieves the same or superior accuracy in the
best hyperparameter. The main bottleneck turned out to be the effect modeling, which is accurate but is too complex for Fast Downward to handle.
Based on this observation, one question arises:
\emph{Can the translator bottleneck be addressed just by using a different rule-based learning method,
such as MAX-SAT based approaches \cite{YangWJ07} or planning based approaches \cite{aineto2018learning}?}
We argue that this is not the case because (1) our Random Forest based DSAMA approach can be considered as the upper bound of
existing Action Model Acquisition method in terms of \emph{accuracy} and (2) should the same accuracy be achieved by other approaches,
\emph{the resulting PDDL must have the same complexity}. We explain the reasoning below.

First, we note that the translation failure is due to the heavy use of disjunctions in the PDDL file for the compilation of Random Forest because, in Fast Downward,
disjunctions are ``flattened'' \cite{helmert2009concise}, i.e., compiled away by making the separate actions for each branch of the disjunction.
This causes an exponential blowup when a huge number of disjunctions are presented to the translator, which is exactly the case for our scenario.
The use of effect conditions are not an issue because Fast Downward uses them directly.

Next, in order to avoid this exponential blowup, the resulting rules learned by the binary classifier must be disjunction-free.
In fact, existing approaches \cite{YangWJ07,aineto2018learning} learn the disjunction-free action models.
One trivial approach to achieve this in DSAMA is to compile a decision tree into Decision Lists \cite{cohen1995fast},
the degenerate case of decision trees where the children (\textit{left}, \textit{right}) of every decision node can contain at most one decision node.
However this is trivially ineffective because compiling a decision tree into a decision list is equivalent to
how Fast Downward makes the actions disjunction-free by splitting them.
Both cases end up in an exponentially large list of disjunction-free actions.

Finally, 
given that our Random Forest based DSAMA achieved almost-perfect accuracy in the successor generation task (effect condition),
we could argue that the rules generated by our approach are quite close to the ground truth rules and, therefore,
\emph{the ground truth rules are at least as complex as the rules found by DSAMA}.
Therefore, if the existing approaches achieved the same accuracy on the same task,
their resulting disjunction-free set of conditions would be as large and complex as the exponentially large ``flattened'' form of our rules.
This argument also applies to the variants of DSAMA using Decision List based classifiers
(e.g., \cite[OneR]{holte1993very},\cite[RIPPER]{cohen1995fast},\cite[MLIC]{maliotov2018mlic}).

\section{Related Work}
\label{sec:related}

Traditionally, symbolic action learners tend to require a certain type
of human domain knowledge and have been situating itself merely as an
additional assistance tool for humans, rather than a system that builds knowledge from the scratch, e.g., from unstructured images.
Many systems require a structured input representation (i.e., First Order Logic) that are partially hand-crafted
and exploits the symmetry and the structures provided by the structured representation,
although the requirements of the systems may vary.
For example,
some systems require state sequences \cite{YangWJ07}, while others require action sequences \cite{CresswellG11,CresswellMW13}.
Some supports the noisy input \cite{MouraoZPS12,zhuo2013action},
partial observations in a state
and missing state/actions in a plan trace \cite{aineto2018learning}, or
a disordered plan trace \cite{zhuo2019learning}.
Approach-wise, they can be grouped into 3 categories:
MAX-SAT based approaches \cite{YangWJ07,zhuo2013action,zhuo2019learning},
Object-centric approaches \cite{CresswellG11,CresswellMW13} and
learning-as-planning approaches \cite{aineto2018learning}.
AMA$_2$ and DSAMA works on a factored but non-structured propositional representation.
While we do not address the problem of lifting the action description,
combining these approaches with the FOL symbols (relations/predicates) found by NN \cite{Asai2019b}
is an interesting avenue for future work.

There are several lines of work that extracts a PDDL action model from a
natural language corpus. Framer \cite{lindsay2017framer} uses a CoreNLP
language model while EASDRL \cite{feng2018extracting} uses Deep
Reinforcement Learning \cite{dqn}. The difference from our approach is
that they are reusing the symbols found in the corpus while we generate
the discrete propositional symbols from the visual perception 
which completely lacks such a predefined set of discrete symbols.

While there are recent efforts in handling the complex state space without
having the action description 
\cite{frances2017purely}, action models could be used for other purposes,
including Goal Recognition \cite{ramirez2009plan},
macro-action generation \cite{BoteaB2015,ChrpaVM15},
or plan optimization \cite{chrpa2015exploiting}.

There are three lines of work that
learn the binary representation of the raw environment.
Latplan SAE \cite{Asai2018} uses the Gumbel-Softmax VAE \cite{MaddisonMT17,jang2016categorical}
which was modified from the original to maximize the KL divergence term for the Bernoulli distribution \cite{Asai2019a}.
Causal InfoGAN \cite{kurutach2018learning} uses GAN\cite{goodfellow2014generative}-based approach
combined with Gumbel Softmax prior and Mutual Information prior.
Mutual Information and the negated KL term are both the same entropy term $H(z|x)$,
i.e., the randomness of the latent vector $z$ given a particular input image $x$.
Latplan ZSAE \cite{Asai2019a} additionally penalizes the ``true'' category in the
binary categorical distribution to suppress the chance of random flips in the latent vector
caused by the input noise. It was shown that these random flips negatively affect the performance
of the recipient symbolic systems by violating the uniqueness assumption of the representation,
dubbed as ``symbol stability problem''.
Quantized Bottleneck Network \cite{koul2018learning} uses quantized activations (i.e., step functions) in the latent space
to obtain the discrete representation. It trains the network with Straight-Through gradient estimator \cite{bengio2013estimating},
which enables the backpropagation through the step function.
There are more complex variations such as VQVAE \cite{van2017neural}, DVAE++\cite{vahdat2018dvae++}, DVAE\# \cite{vahdat2018dvae}.

In the context of modern machine learning,
Deep Reinforcement Learning (DRL) has solved complex problems,
including Atari video games \cite[DQN]{dqn} or Game of Go \cite[AlphaGo]{alphago}.
However, they both have a hard-coded list of action symbols (e.g., levers, Fire button, grids to put stones)
and relies on the hard-coded simulator for both learning and the correct execution.

In another line of work, Neural Networks model the
external environment captured by video cameras by
explicitly taking the temporal dependency into account \cite{lotter2016deep}, unlike Latplan SAE,
which processes each image frame one by one.

\section{Conclusion}

In this paper, we negatively answered a question of \emph{whether simply replacing a
neural, black-box Action Model Acquisition model with a rule-based machine learning model
would generate a useful descriptive action model from the raw, unstructured input.}
Our approach hybrids a neural unsupervised learning approach to the action label generation
and the precondition/effect-condition learning using \sota rule-based machine learning.
While the proposed method was able to generate accurate PDDL models,
the models are too complex for the standard planner to preprocess
in a reasonable runtime and memory.

The fact that the rather straightforward modeling of effects in DSAMA is causing such a huge problem is worth noting.
The planning domains written by humans, in general, tend to have a specific human-originated property that causes this type of
phenomenon to happen less often, and this might be reflected by the fact that STRIPS (without disjunctions) was the first common
language adapted by the community.
Unlike the planning models written by the human, we found that the set of propositions generated by the State AutoEncoder
network, as well as the set of action labels generated by the Action
AutoEncoder network, do not have such a property.
The state space and the clustering of transitions are ``less organized'' compared to the typical human models, and
the lack of human-like regularization behavior makes an otherwise trivial task of PDDL-SAS+ translation intractable in a modern planner.

The future directions are twofold. The first one is to
find the right regularization or the right architecture for the neural networks in order to
further constrain the space of the ground-truth transition model in the latent space.
This is similar to the approach pursued by \cite{Asai2019a}
which tries to suppress the instability of the propositional values in the latent space.
Machine Learning community is increasingly focusing on the disentangled representation learning \cite{higgins2017beta}
that tries to separate the meaning of the feature values in the latent space.
Finding the right structural bias for neural networks has a long history,
notably the convolutional neural networks \cite{fukushima1980neocognitron,lecun1989backpropagation,krizhevsky2012imagenet} for images, or
LSTMs \cite{hochreiter1997long} and transformers \cite{vaswani2017attention} for sequence modeling.

The second approach is to develop a planner that can directly handle the complex logical conditions in an efficient manner.
Fast downward requires converting the input PDDL into SAS+ with a rather slow translator,
assuming that such a task tends to be easy and tractable. While this may hold for most hand-crafted domains (such as IPC domains),
it may not be a viable approach when the symbolic input is generated by neural networks.

\fontsize{9.5pt}{10.5pt}
\selectfont

\end{document}